\documentclass[conference,letterpaper]{IEEEtran}
\IEEEoverridecommandlockouts
\usepackage{graphicx}  % Written by David Carlisle and Sebastian Rahtz
\usepackage{multirow}
\usepackage{algorithmic}
\usepackage{algorithm}
\usepackage{times}
\usepackage{cite}
\usepackage{epsfig}
\usepackage{graphicx}
\usepackage{lipsum}
\usepackage{amsmath}
\usepackage{amssymb}
\usepackage{listings}
\usepackage[left=0.71in,top=0.94in,right=0.71in,bottom=1.18in]{geometry}
\begin{document}
% paper title
\title{\huge Enhancement Mask for Hippocampus Detection and Segmentation}

% author names and affiliations
 \author{\authorblockN{Dengsheng Chen, Wenxi Liu, You Huang, Yuanlong Yu}
 \authorblockA{\textit{College of Mathematics and Computer Science}\\
 \textit{Fuzhou University}\\
}%
 \and
 \authorblockN{Tong Tong}
 \authorblockA{\textit{Imperial Vision}\\
 \\}}%

% make the title area
\maketitle
\begin{abstract}
  Detection and segmentation of the hippocampal structures in volumetric brain images is 
  a challenging problem in the area of medical imaging. In this paper, we propose
  a two-stage 3D fully convolutional neural network that efficiently detects and segments the 
  hippocampal structures. In particular, our approach first localizes the hippocampus 
  from the whole volumetric image while obtaining a proposal for a rough segmentation. 
  After localization, we apply the proposal as an enhancement mask to extract the fine structure of the hippocampus. 
  The proposed method has been evaluated on a public dataset and compares with state-of-the-art approaches.
  Results indicate the effectiveness of the proposed method, which yields mean Dice Similarity Coefficients
  (i.e. DSC) of $0.897$ and $0.900$ for the left and right hippocampus, respectively. Furthermore, 
  extensive experiments manifest that the proposed enhancement mask layer has remarkable benefits
  for accelerating training process and obtaining more accurate segmentation results.

\end{abstract}

% key words

\section{Introduction}

With pervasive applications of medical imaging, biological structure detection and segmentation have
been fundamental and crucial tasks in biomedical imaging research. It extracts different tissues, 
organs, pathologies and biological structures, to support medical diagnosis surgical planning and treatments. 
In practice, the detection and segmentation are performed manually by pathologists, 
which is time-consuming and tedious. The ever-increasing variety of medical images make manual segmentation 
impracticable in terms of cost and reproducibility.
Thus, automatic biomedical detection and segmentation are highly desirable. 
However, this task is extremely challenging, because of the 
heterogeneous of biological objects including the large variability in location, size, shape and 
frequency, and also because of low contrast, noise and other imaging artifacts caused by various medical
imaging moralities and techniques \cite{Milletari2016V}.

In the past years, substantial progress has been made in biomedical image detection and segmentation with pixel-based methods 
\cite{Doyle2006A,Nguyen2012Structure,Tabesh2007Multifeature,Sirinukunwattana2015A}and structure-based methods 
\cite{Altunbay2010Color,Gunduz2010Automatic,Fu2014A,Sirinukunwattana2015A_}. 
In recent years, the fully convolutional networks (FCNs \cite{Long2017Fully}), 
have been used for biomedical image segmentation, which require little hand-crafted features or prior knowledge. 
FCNs trained end-to-end have been previously applied to 2D images both in computer
vision \cite{Noh2015Learning} and microscopy image analysis \cite{Ronneberger2017Invited}. 
These models are trained to predict a segmentation mask, delineating the structures of interest, for the whole image. 
Ronneberger et al. \cite{Ronneberger2017Invited} proposed U-Net, a deep convolutional network that adds skip connections 
to the symmetric feature maps to perform 2D medical image segmentation. With data augmentation, 
it achieves significant improvement over previous methods. 
Recently, this approach was extended to 3D and applied to segmentation of volumetric data acquired from
a confocal microscope \cite{o20163D}. With the inspiration of these models, 
Fausto et al. \cite{Milletari2016V} divided model into stages that learn residuals and as empirically 
observed improve both results and convergence time. 
However, due to the heavy computational burden of 3D convolutional operation and a large number 
of uncertain parameters, it is formidable to devise an extremely deeper 3D neural network which inherits more hidden features. 
Furthermore, with the limitation of hardware devices, the prior methods have to downsample the high resolution medical 
image before feeding into networks, which gives rise to the loss of segmentation accuracy. 

To handle the problems mentioned above, we present a novel two-stage 3D fully convolutional neural network,
for detecting and segmenting biomedical objects from volumetric medical images, e.g. MRI. Our approach not only reduces the 
computational burden of 3D FCN, but also preserve the segmentation details. 
In particular, our model first learns the localization and segmentation context of the biomedical objects from the 
low resolution input and then integrate it with the original input by an enhancement mask that performs
the fine structure segmentation. We experiment the proposed approach in a hippocampus segmentation dataset (ADNI) 
and achieve state-of-the-art results.

\noindent\textbf{Main Contributions: }
We present a two-stage 3D fully convolutional neural network that efficiently detects and then segments the hippocampal 
structure from volumetric brain images. Our framework first learns the holistic structural information from the downsampled 
input image, which localizes the hippocampus, while roughly estimating the segmentation through a proposal network. 
In the second stage, we introduce an enhancement mask that integrates the rough segmentation proposal in a segmentation 
network to perform fine segmentation. 

\begin{figure}
\begin{center}
\includegraphics[width=3.2in]{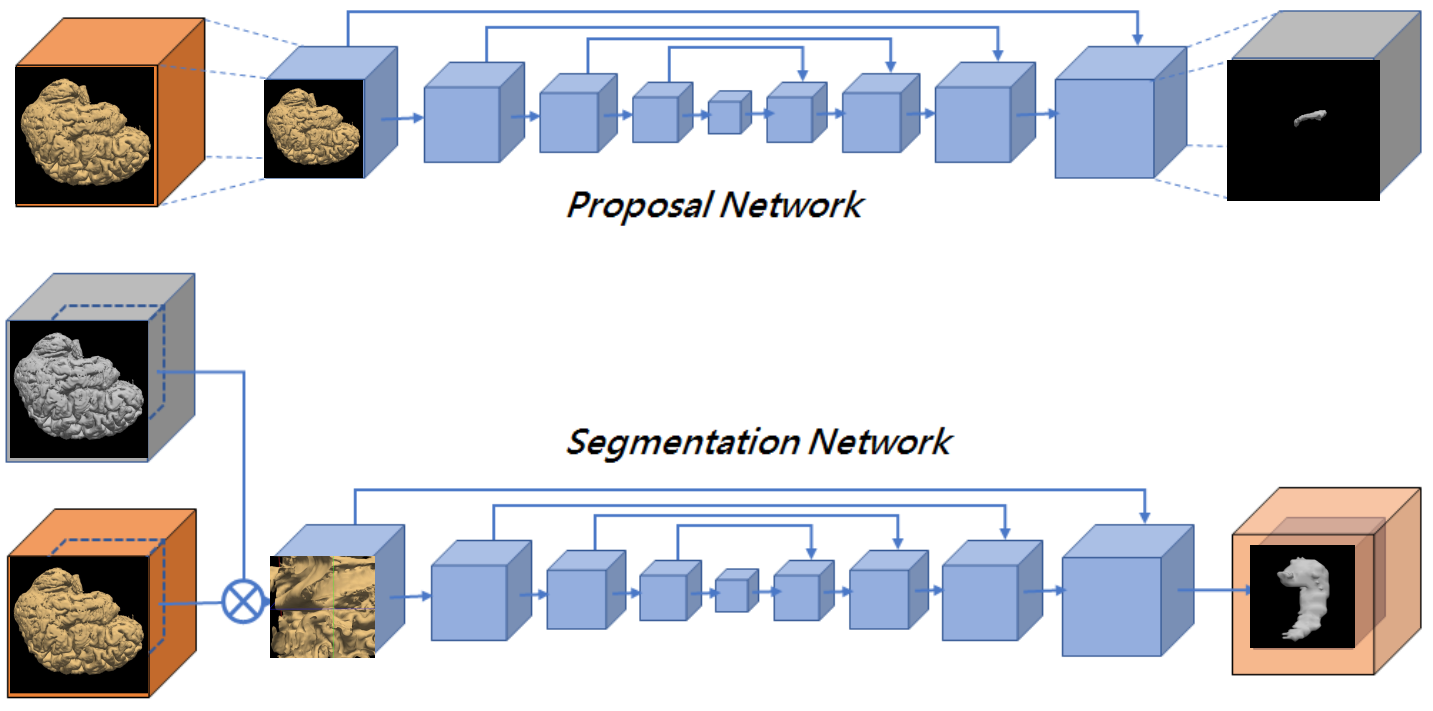}
\end{center}
  \caption{Our framework consists of two-stage networks. 
  In the first stage, the input data (i.e. the orange cube) is downsampled and fed into a 3D fully convolutional neural network, 
  called \textit{Proposal Network} which learns a holistic probability map. An enhancement mask will be generated from this probability
  map. The enhancement mask will be applied to the original data. We will crop a $64 \times 64 \times 64$ cube according to 
  hippocampus localization from the enhancement data as the input of \textit{Segmentation Network}. Hence, the Segmentation network outputs 
  the fine segmentation result of the cropped data.
  }
\label{fig:long}
\label{fig:onecol}
\end{figure}

\section{The Proposed Method}

In this paper, we propose a two-stage framework to detect the hippocampus from the 
volumetric brain images and then apply segmentation to the enhanced original image data through 
the enhancement mask layer. Our proposed framework can fully exploit the holistic information of 
the volumetric data and also efficiently perform the segmentation while preserving the fine structure 
of the hippocampus. Specifically, in the first stage (\textit{Proposal Network}), 
the hippocampus is localized from the downsampled input volumetric data while a rough segmentation proposal
of the hippocampus is learned with a 3D FCN. In the second stage (\textit{Segmentation Network}), 
the original input data is enhanced through the enhancement mask which generates from \textit{Proposal Network} and then cropped.
We will introduce these two stages in the following subsections.

\subsection{\textit{Proposal Network} and Localizing Hippocampus}

In our application, each image consists of several slices with the resolution of $256\times256$
(pixels) and the numbers of slices range from 166 to 256. Processing the original volumetric data 
requires a large amount of memory on the runtime. However, the holistic view of the 
volumetric data provide the crucial context to localize and segment the hippocampus. 
To fully exploit the holistic information of the input volumetric data, we build up 
\textit{Proposal Network} to estimate the segmentation of the hippocampus. The architecture of \textit{Proposal Network} 
is an encoder-decoder, which learns a probability map about pixelwise.
As shown in Fig.1, \textit{Proposal Network} first compress the input data via multiple 
convolutional layers into feature maps with smaller resolutions, and then perform 
decompression by upsampling the feature maps layer-by-layer via de-convolution 
and combining with corresponding compressed feature maps respectively by skip-connection. 
The output of the network is the response of the segmented hippocampus which performs
the pixelwise estimation on whether it belongs to the hippocampus. 

To learn the holistic information, the original input data $\textbf{X}$ is downsampled to 
$64 \times 64 \times 128$ and fed into \textit{Proposal Network} $\mathcal{F}$. 
At the end, the output response map is upsampled to the original size using 3D bilinear interpolation
sampling technique. 
The whole procedure is denoted as $\mathcal{F}(\textbf{X}) \in \mathbb{R}^{W\times H \times D}$.
Intuitively, in the proposal network, learning the segmentation response from the downsampled
low-resolution data actually enlarges the receptive field to cover the entire volumetric image.
Although the accuracy of the segmentation at this stage is not satisfactory, the output response
provides the crucial context for localizing the hippocampus whose volume is much smaller compared 
to the whole brain, and the rough segmentation results still make a vital improvement on further
image segmentation through the enhancement mask which is generated from itself. Furthermore, since the whole image data is fed into \textit{Proposal Network}, it makes the detection can be extended to other segmentation problems.

\noindent {\bf Localizing Hippocampus} Before the segmentation, we need to localize the hippocampus in advance. 
In essence, the proposal response $\mathcal{F}(\textbf{X})$ 
indicates the probability of whether the hippocampus shows up in each voxel. We denote it as $\textbf{L}_{proposal}$ for clarity. 
The goal is to calculate the central coordinate of the hippocampus from $\textbf{L}_{proposal}$.
Instead of using sliding window or clustering, we accumulate the voxel value of $\textbf{L}_{proposal}$ which is a tensor with the size of $W\times H\times D$ along the X, Y, and Z axis respectively. Taking the X axis for instance, the accumulation is computed as follows: $\textbf{H}_X(i) = \sum_{j=1}^H\sum_{k=1}^D \textbf{L}_{proposal}(i,j,k),\forall i \in [1,W]$. If $\textbf{H}_X(i)$ is larger than a certain threshold $\epsilon$, it indicates the occurrence of the hippocampus at the $i^{th}$ slice along the X axis. 
Empirically, we set $\epsilon$ as $5$ and it can effectively and efficiently filter out the false alarm in $\textbf{L}_{proposal}$. Hence, we can compute the boundary coordinates by: 
\begin{align}
    &X_{min} = \arg\min_i \textbf{H}_X(i), \\\nonumber &X_{max} = \arg\max_i \textbf{H}_X(i) \\ \nonumber
    &Y_{min} = \arg\min_i \textbf{H}_Y(i), \\\nonumber &Y_{max} = \arg\max_i \textbf{H}_Y(i) \\ \nonumber
    &Z_{min} = \arg\min_i \textbf{H}_Z(i), \\\nonumber &Z_{max} = \arg\max_i \textbf{H}_Z(i) \\
    s.t. & \text{   } \textbf{H}_X(i)>\epsilon, \textbf{H}_Y(i)>\epsilon, \textbf{H}_Z(i)>\epsilon
\end{align}
% Note that there are a pair of hippocampal structures in the brain image and the proposal response $\mathcal{F}(\textbf{X})$ may contain scales of noise. We simply filter out the small clusters by eliminating the cluster with less than 5 voxels. Hence, the two largest clusters correspond to the left and right hippocampus. 
% To find the optimal cuboid volume that contains the clusters, we just need to compute the geometric center.
The central coordinate of localizing the hippocampus is computed:  $\textbf{V}_{center}=\{(X_{max}+X_{min})/2, (Y_{max}+Y_{min})/2, (Z_{max}+Z_{min})/2\}$, where $X_{max}$, $X_{min}$, $Y_{max}$, $Y_{min}$, $Z_{max}$, $Z_{min}$ are the maximum and minimum coordinates along all axes. 

\subsection{Enhancement Mask Layer and \textit{Segmentation Network}}

\noindent {\bf Enhancement Mask Layer}
The enhancement mask \textbf{M} is generated from the rough segmentation results of \textit{Proposal 
Network} by the following formulation:
\begin{equation}
	\textbf{M} = \textbf{L}_{proposal} * \alpha + \beta
\end{equation}
where the $\textbf{L}_{proposal}$ contains the holistic and reliable information which 
regard to the location of the hippocampus, but the details of its 3D structure is blurred 
due to resizing which needs to be compensated by the original data $\textbf{C}_{original}$. Both $\alpha$, 
$\beta$ are the parameters which control the influence of the enhancement mask on the 
original image data. It is obviously that the bigger $\alpha$ and smaller $\beta$ will cause a 
stronger influence of enhancement mask. As for $\textbf{L}_{proposal}$ is 0/1 label value, we set
$\beta$ as 1 which respect for that we do not apply any enhance or suppress operation on the original 
background image data. The value of $\alpha$ is relatively flexible, and different $\alpha$ value will
cause a significant different performance on the segmentation networks. A detail 
analysis on how to select a suitable $\alpha$ value will be shown in next section.

\noindent {\bf Segmentation Network}
In order to preserve the detailed structure of the hippocampus in segmentation, 
it is common practice to empirically crop a portion of voxels from the original volumetric image and 
then infer the segmentation based on this cropped volume. Although such cropping operation keeps the 
nearing-range context around the hippocampus while reducing the computation burden, it abandons most 
of the long-range dependent spatial information in the whole brain structure, which bounds the 
segmentation performance. What's worse, the empirically cropping operation does not guarantee for dealing 
with the data error caused by medical instrument malfunction, and we have to adapted the crop localization
according to different objects which is quietly inconvenient in practice. 
To introduce the holistic view of the original input data, we need to incorporate the previously inferred 
enhancement mask $\textbf{M}$ that contains the holistic information with the original data as follows: 
\begin{equation}
	\textbf{X}^\prime = \textbf{M} \odot \textbf{C}_{original},
\end{equation}
\noindent where $\textbf{X}^\prime$ ($\textbf{X}^\prime\in \mathbb{R}^{w\times h \times d}$) is the enhanced 
input data that will be fed into the segmentation network and $\odot$ denotes the dot-product operator. 

Then we crop a volume with the size of $h \times w \times d$ that covers the target region and centering at 
$\textbf{V}_{center}$.

Given the estimated location of the hippocampus,
we also need to incorporate the holistic information to the cropped data before feeding into the segmentation 
network. We denote the cropped $h \times w \times d$ volume centered at $\textbf{v}_{center}$ 
of the probability map $\mathcal{F}(\textbf{X})$ as $\textbf{C}_{proposal} = Crop(\mathcal{F}(\textbf{X}), 
\textbf{v}_{center})$. We also crop a volume from the exact same region of the original input 3D data 
$\textbf{X}$, which is denoted as $\textbf{C}_{original} = Crop(\textbf{X},\textbf{v}_{center})$. 

On one hand, the proposal serves as the attention mask to enhance the likely regions of the hippocampus
localization to guide the segmentation. On the other hand, adding bias to the proposal prevent from the 
information loss, since it does not completely depress the regions with low responses in proposal. 
Intuitively, the proposal with low responses may still be part of the hippocampus. 
The experiments show that the proposal-enhanced data actually boosts the estimation 
accuracy and speeds up the convergence rate of training the fully convolutional neural network. 

At the end, the enhanced input $\textbf{X}^\prime$ is fed into another network that 
has the same architecture as the proposal network, to create the fine segmentation result. 

\subsection{Training Loss}
The network is trained by \textit{Dice Loss}. The variable Dice
Loss $D$ between two binary segmentation volumes $P$ and $G$ is the same as \cite{Milletari2016V},
written as
\begin{equation}
    D(P, G) = \frac{2\sum_{i=1}^N p_i g_i}{\sum_{i=1}^N p_i^2 + \sum_{i=1}^N g_i^2}
\end{equation}\label{eq:weight}
where the sums run over the $N$ voxels, of the predicted
binary segmentation volume $p_i \in P$ and the ground truth binary volume $g_i \in G$. 
% This formulation of Dice can be differentiated with respect to the j-th voxel of the prediction,
% yielding the gradient:
% \begin{equation}
%   \frac{\partial D}{\partial p_j}
% = 2 \left[ \frac{g_i \left(\sum_{i}^N p_i^2 + \sum_{i}^N g_i^2 \right) - 
%   2 p_j \left(\sum_{i}^N p_i q_i \right)}{\left( \sum_{i}^N p_i^2 + \sum_{i}^N g_i^2\right)^2}
% \right]
% \end{equation}

\section{Experimental Results}

\subsection{Datasets}
Described in this section are several experiments conducted to evaluate the performance of 
our method for campus segmentation using the publicly available ADNI database
\footnote{http://adni.loni.usc.edu/}. The size of the voxels of the image is $1\times 1\times 1 mm^3$.
Each image consists of several slices with the resolution of $256 \times 256$ (pixels) and
the numbers of slices range from 166 to 256. We utilized 110 normal control subjects, 
and downloaded their baseline T1-weighted whole brain MRI images along with their hippocampus masks. 
We train our network as ten-fold cross validation.

\subsection{Implementation details}
\textit{Proposal Network} has the identical structure and parameter settings as 
\textit{Segmentation Network} despite the different input image size. 
Each convolution and de-convolution layer has a $3 \times 3 \times 3$ kernel size, 
and each down and up sampling convolution layer has a $2 \times 2 \times 2$ kernel size. 
At each block of convolutional layer, the output size is half down and the output channel is double.
At each block of de-convolutional layer, the output size is double and the output channel is half 
down on the contrary. The input data of \textit{Proposal Network} is the resized image, while the other is
cropped-masked image. The different weight $\alpha$ and bias $\beta$ will cause significantly 
different performance for segmentation network. Empirically, we set $\beta$ as 1 which means 
the original data will not be suppressed or impressed by $\beta$. In other words, 
dot production with the mask ranging from zero to one will degrade the data. 
Our mask will preserve the original object details. 
The value of $\alpha$ determines the influence of the mask which gained from the proposal network. 
As for that the output of the proposal network is not good enough, the value of $\alpha$ should 
be well designed. In order to explore the optimal weight, we compare different weights in 
Sec.~\ref{sec:weight}. 

\subsection{Evaluations Metric}

To quantitatively evaluate the proposed method, four metrics were used for performance evaluation. 
The degree of overlap was measured for two ROIs $V_s$ and $V_g$, where $V_s$ and $V_g$ are the sets 
of object(hippocampus) voxels automatically segmented by the segmentation method and manually segmented 
by clinical expert, respectively.
\begin{itemize}  
\item {\bf Dice similarity coefficient (DSC)} is a comprehensive similarity metric 
that measures the degree of overlap of two ROIs
$$
    DSC = 2 \times \frac{\arrowvert V_s \cap V_g \arrowvert}{\arrowvert V_s 
    \arrowvert + \arrowvert V_g \arrowvert}
$$
where $\arrowvert \centerdot \arrowvert$ is the cardinality of a set.
\item {\bf Jaccard similarity coefficient (JSC)}, which is a statistic used for comparing 
the similarity and diversity of two ROIs, is described as follows:
$$
    JSC = \frac{\arrowvert V_s \cap V_g \arrowvert}{\arrowvert V_s 
    \cup V_g \arrowvert} = \frac{\arrowvert V_s \cap V_g \arrowvert}{\arrowvert V_s 
    \arrowvert + \arrowvert V_g \arrowvert - \arrowvert V_s \cup V_g \arrowvert}
$$
\item {\bf Precision Index (PI)} is the ratio between the overlap of 
two ROIs and the ROI manually segmented by clinical expert, as follows:
$$
    PI = \frac{\arrowvert V_s \cap V_g \arrowvert}{\arrowvert V_g \arrowvert}
$$
\item {\bf Recall Index (RI)} is the ratio between the overlap of two ROIs 
and the ROI segmented by the segmentation method, as follows:
$$
    RI = \frac{\arrowvert V_s \cap V_g \arrowvert}{\arrowvert V_s \arrowvert}
$$

\end{itemize}  

A larger value of all the metric mentioned above indicate a better segmentation 
performance.

\subsection{Selecting weights for enhancement mask}
\label{sec:weight}
To select optimal value of weights for the mask, we need to analyze the effects of the weight $\alpha$ in training and testing in the first place. 

\begin{figure}
\begin{center}
\includegraphics[width=3.2in]{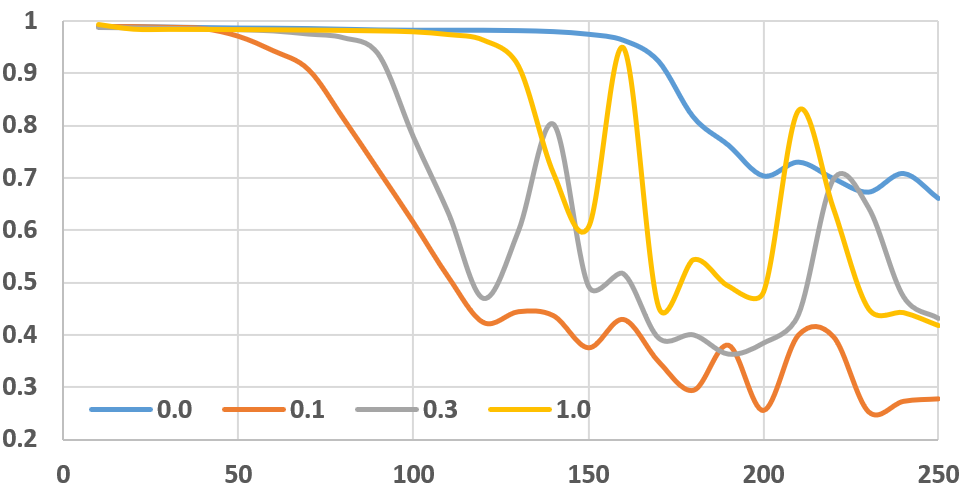}
\end{center}
  \caption{We illustrate the training loss(\textbf{Dice Loss}) of our model with varied mask weights,
   i.e., $\alpha=\{0.0, 0.1, 0.3, 1.0\}$, for the first $250$ iterations.}
\label{fig:train}
\end{figure}

\begin{figure}
\begin{center}
\includegraphics[width=3.2in]{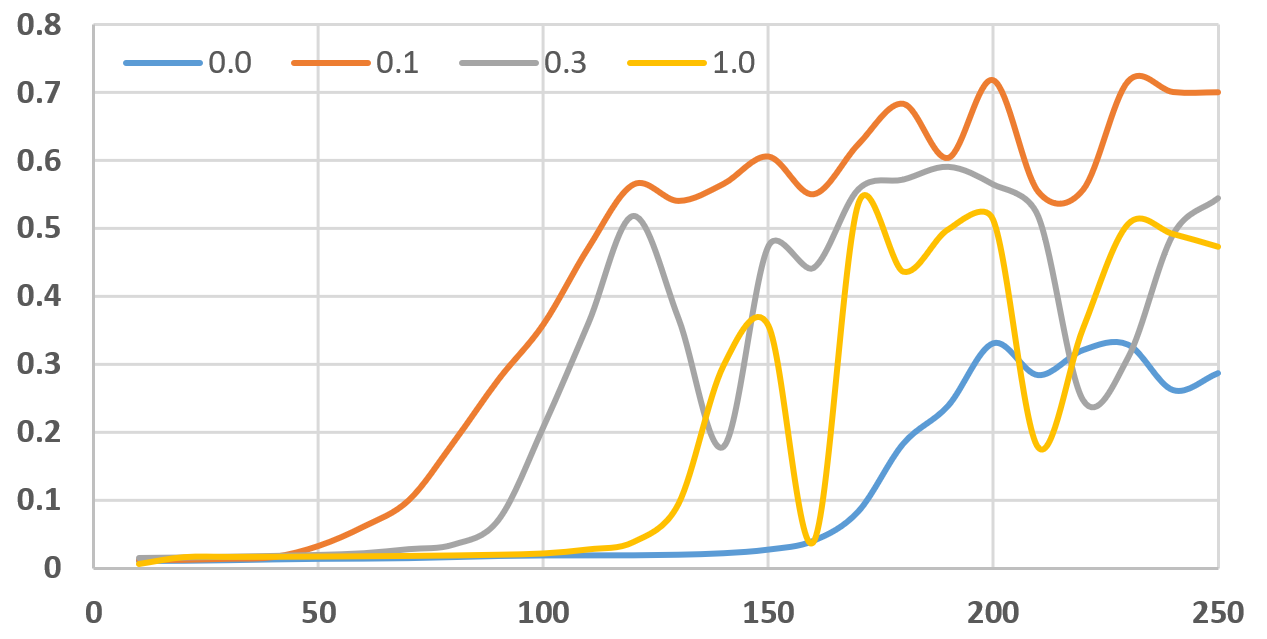}
\end{center}
  \caption{We illustrate the testing \textbf{DSC} of our model with varied mask weights, 
  i.e., $\alpha=\{0.0, 0.1, 0.3, 1.0\}$, for the first $250$ iterations.}
\label{fig:dsc}
\end{figure}

To do so, we assign different values to $\alpha$, i.e., $0, 0.1, 0.3, 1.0$. We first evaluate how the weight influences the training process. 
We illustrate the training loss (i.e. Dice Loss) in the first 250 epochs with different values of $\alpha$. According to Fig.~\ref{fig:train}, 
as $\alpha=0.1$, the training loss drops quickly. With $\alpha=0.3$ and $1.0$, the loss also decreases quickly but their curves are not stable. 
Comparably, without mask (i.e. $\alpha=0$), the training appears to be slow. 
It indicates that using the enhancement mask will easily speed up the training, but it may also have the chance to make the training unstable. 
Besides, we also evaluate the same models with different $\alpha$ in the validation dataset. 
Their performance is measured using DSC as shown in Fig.~\ref{fig:dsc}. Similar to the observation above, 
the model with $\alpha=0.1$ has the optimal performance in testing. 
With $\alpha=0.3$ and $1.0$, the DSC curves are unstable. 
To sum up, $\alpha$ lets the mask have more impact on the input data, which speeds up the training and 
enhance the performance of the model. However, too large weights may also cause the overfitting of the trained models. 
Among these weights, $\alpha=0.1$ allows the mask to shed some light on the segmentation without jeopardizing the training model.
Hence, in the following experiments, we choose $\alpha=0.1$ as the weight of the enhancement mask.

% (Figure: training loss curve)

\begin{figure}
\begin{center}
\includegraphics[width=3.2in]{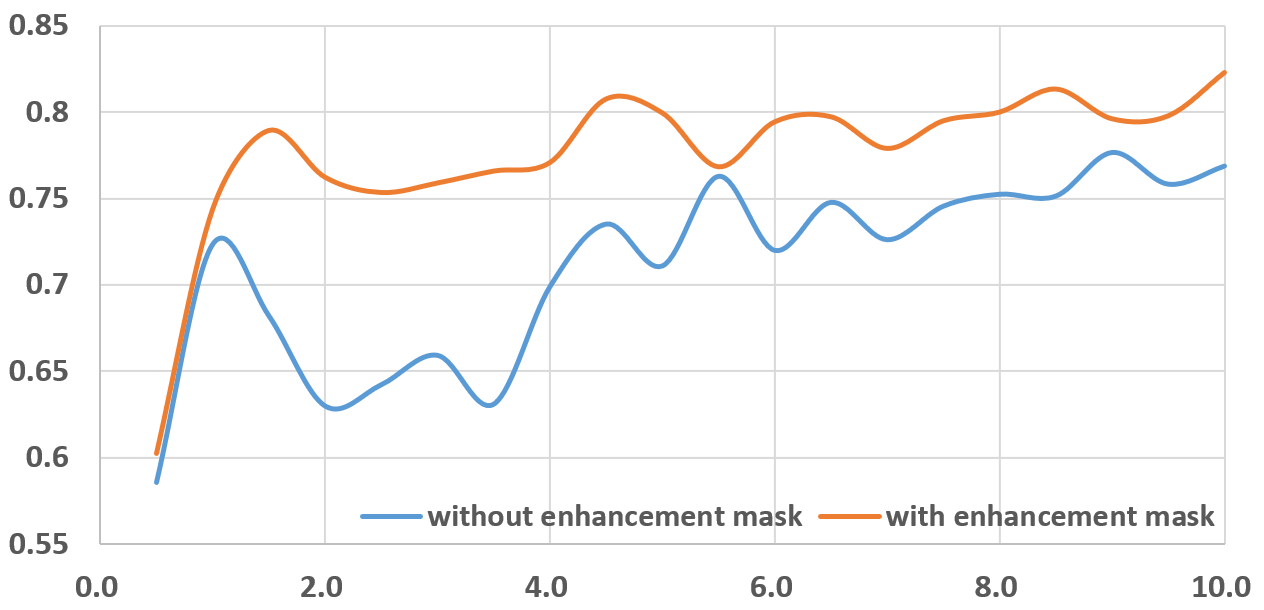}
\end{center}
  \caption{Comparison between $\alpha = 0.0$ (without mask) and $\alpha = 0.1$ after training for $10K$ iterations.}
\label{fig:10k}
\end{figure}

\subsection{Ablation study}

To demonstrate the effect of the mask, we perform comprehensive experiments. 
First, as shown in Fig.~\ref{fig:10k}, we compare the DSC for the models with and without enhancement mask after training for $10,000$ 
iterations and evaluate their segmentation performance every $500$ iterations in the validation dataset. 
It illustrates that the model with mask achieves significantly better performance after long-term training.
Second, as shown in Tab.~\ref{tab:combo}, we train the proposal network with iterations ranging from $500$ to $5,500$, respectively. 
They are denoted as $\mathrm{N}^{0.5}_{pr}$, $\mathrm{N}^{1.0}_{pr}$,$\dots$,$\mathrm{N}^{5.5}_{pr}$. 
Similarly, we train the segmentation network 
with different numbers of iterations as well. They are denoted as $\mathrm{N}^{0.0}_{seg}$, 
$\mathrm{N}^{0.5}_{seg}$,$\dots$, $\mathrm{N}^{5.0}_{seg}$. 
Generally, the performance of networks gets better after more training epochs, e.g., $\mathrm{N}^{2.5}_{pr}$ 
is better than $\mathrm{N}^{2.0}_{pr}$, and $\mathrm{N}^{4.0}_{seg}$ is worse than $\mathrm{N}^{5.0}_{seg}$. 
We pair each proposal network, $\mathrm{N}^{i}_{pr}$, with a segmentation network, $\mathrm{N}^{j}_{seg}$, 
to form a variant of our model. We evaluate these variants to analyze how the proposal network and the segmentation network 
affect each other. Observing columns in Tab.~\ref{tab:combo},
using better proposal network with the same segmentation network may lead to a boost of the performance, 
especially combining with better segmentation network. It turns out the proposal indeed makes contributions in the segmentation.
\begin{figure}
\begin{center}
\includegraphics[width=3.2in]{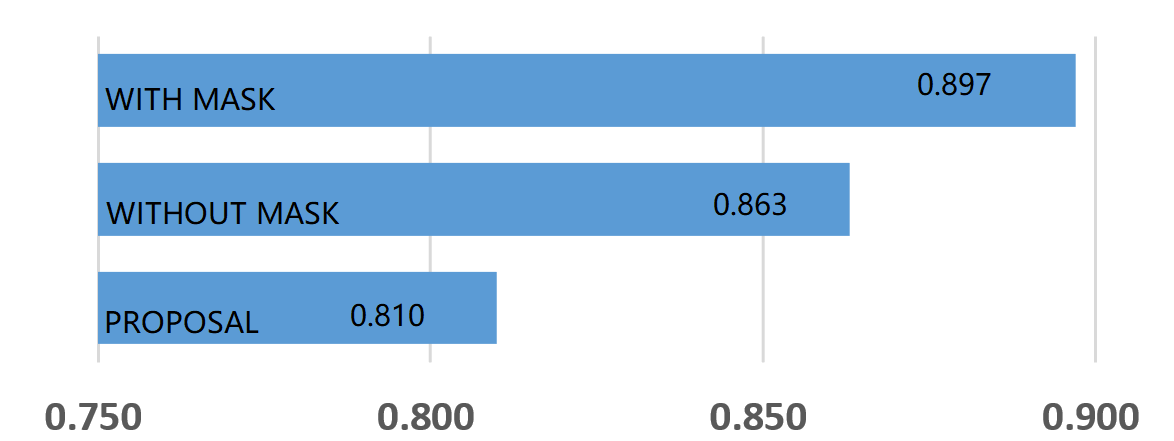}
\end{center}
  \caption{Comparison of \textit{proposal network}, 
  \textit{segmentation network without enhancement mask} and \textit{
  segmentation network with enhancement mask (\textbf{DSC})}.}
\label{fig:self_comp}
\end{figure}
 Another fact is that the model can achieve satisfactory performance without fully trained proposal network. 
 For instance, after training the proposal network for only $1500$ epochs, the model can reach more than $0.86$ accuracy in DSC 
 that is comparable to the state-of-the-art approaches. This is another benefit of the enhancement mask, which give our confidence
 for reducing the training iterations of \textit{Proposal Network} to less than 1000.
% (Table of pairs of proposal network and segmentation network)
\begin{table*}[]
\footnotesize
\centering
\caption{Combination of \textit{Proposal Network} and \textit{Segmentation Network} trained with different numbers of iterations.
        The \textit{0.0 K Iter.} in the first column indicates the original segmentation DSC of \textit{Proposal Network}.}
\label{tab:combo}
\begin{tabular}{c|ccccccccccc}
\hline
\textbf{Iteration (\textit{K})} & \textbf{0.0} & \textbf{0.5} & \textbf{1.0} & \textbf{1.5} & \textbf{2.0} & \textbf{2.5} & \textbf{3.0} & \textbf{3.5} & \textbf{4.0} & \textbf{4.5} & \textbf{5.0} \\ \hline
\textbf{0.5} & 0.5856       & 0.6299       & 0.7458       & 0.7726       & 0.7570       & 0.7510       & 0.7498       & 0.7550       & 0.7596       & 0.7861       & 0.7796       \\
\textbf{1.0} & 0.7232       & 0.7397       & 0.8248       & 0.8408       & 0.8363       & 0.8406       & 0.8433       & 0.8457       & 0.8473       & 0.8540       & 0.8543       \\
\textbf{1.5} & 0.6818       & 0.7529       & 0.8377       & 0.8534       & 0.8487       & 0.8498       & 0.8562       & 0.8560       & 0.8573       & \textbf{0.8649}       & 0.8664       \\
\textbf{2.0} & 0.6297       & 0.7596       & 0.8407       & 0.8538       & 0.8500       & 0.8514       & 0.8559       & 0.8562       & 0.8580       & 0.8645       & 0.8655       \\
\textbf{2.5} & 0.6421       & 0.7515       & 0.8366       & 0.8516       & 0.8456       & 0.8458       & 0.8525       & 0.8535       & 0.8551       & 0.8631       & 0.8629       \\
\textbf{3.0} & 0.6590       & 0.7516       & 0.8356       & 0.8523       & 0.8464       & 0.8462       & 0.8531       & 0.8530       & 0.8551       & 0.8636       & 0.8633       \\
\textbf{3.5} & 0.6308       & 0.7612       & 0.8413       & 0.8546       & 0.8516       & 0.8520       & 0.8571       & 0.8573       & 0.8588       & 0.8652       & 0.8652       \\
\textbf{4.0} & 0.6988       & 0.7504       & 0.8339       & 0.8478       & 0.8439       & 0.8464       & 0.8518       & 0.8520       & 0.8537       & 0.8618       & 0.8607       \\
\textbf{4.5} & 0.7349       & 0.7578       & 0.8408       & 0.8563       & 0.8533       & 0.8547       & 0.8589       & 0.8607       & 0.8617       & 0.8674       & 0.8687       \\
\textbf{5.0} & 0.7107       & 0.7542       & 0.8372       & 0.8518       & 0.8489       & 0.8500       & 0.8548       & 0.8549       & 0.8573       & 0.8643       & 0.8637       \\
\textbf{5.5} & 0.7625       & 0.7648       & 0.8472       & 0.8610       & 0.8595       & 0.8607       & 0.8647       & 0.8659       & 0.8659       & 0.8708       & \textbf{0.8728}      \\ \hline
\end{tabular}
\end{table*}
Finally, we evaluate the proposal network with the segmentation networks with and without mask in the validation dataset. 
As we observe in Fig.~\ref{fig:self_comp}, the proposal network has only 0.810 accuracy in DSC and the segmentation network 
without mask gain 0.863. The model with the enhancement mask performs better than them with 0.897. 
As an example shown in Fig.~\ref{fig:result}, the top-left figure shows the manual labeled hippocampus. 
The top-right figure shows the output proposal of the proposal network with the original resolution. 
It correctly localizes the hippocampus, but it is difficult to discriminate the structural details. 
The bottom two figures visualize the hippocampus segmentation with and without mask. 
As illustrated in the area pointed by the red arrow, the result generated by the model with mask has fewer 
artifacts and more accurate segmentation in the tail of the hippocampus.
\begin{figure}
\begin{center}
\includegraphics[width=3.2in]{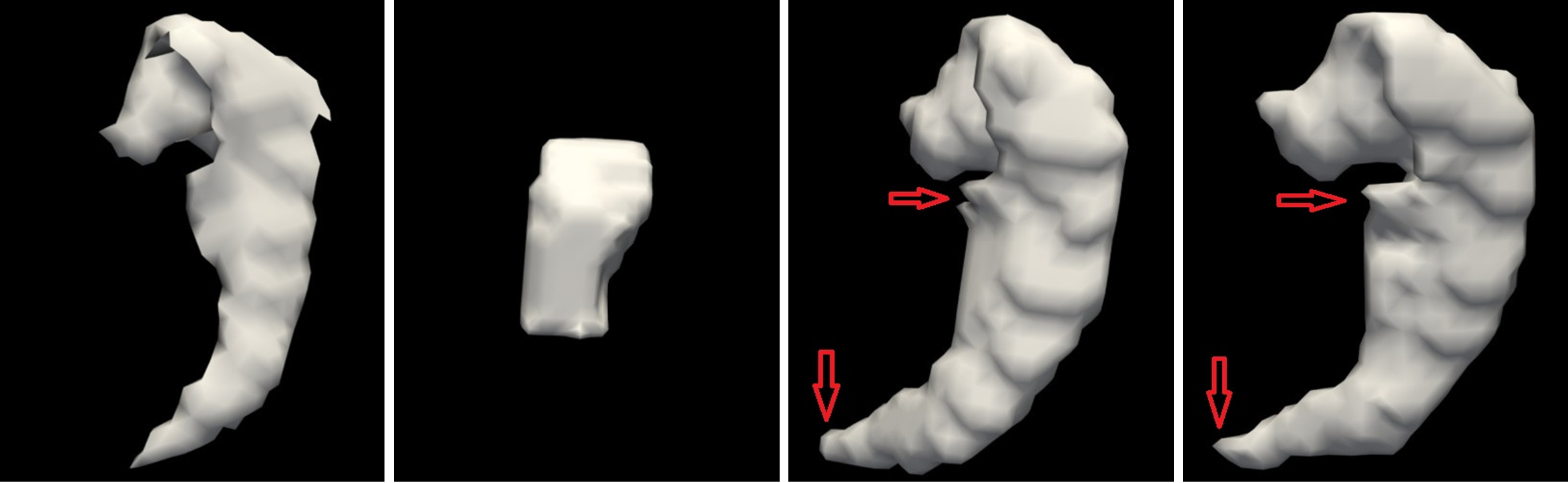}
\end{center}
  \caption{Hippocampal detection and segmentation results by our method.
  \textbf{first}: the manual segmentation hippocampal structure. 
  \textbf{second}: the results of our \textit{Proposal Network}.
  \textbf{third}: the segmentation of the hippocampal structure without 
                enhancement mask.
  \textbf{fourth}: the segmentation of the hippocampal structure with 
                enhancement mask.}
\label{fig:result}
\end{figure}
\subsection{Quantitative comparison}

In this experiment, we adopt 10-fold cross-validation strategy to evaluate the segmentation of the left and the right hippocampus respectively.
For comparison purposes, the conventional patch-based method by non-local weighting (Nonlocal-PBM)\cite{Coup2010Nonlocal}, the recently proposed sparse patch-based labeling (Sparse-PBM)\cite{Zhang2012Sparse} and Path-Based Label Fusion (PBLF)\cite{Wang2018Patch} are evaluated on the samples collected on ADNI dataset. 

We compare the proposed model with these methods and 
evaluate on four metrics as shown in Tab.~\ref{tab:comparison}. 
According to the quantitative comparison, our approach generally outperforms other methods with the obvious advantages in the metrics DSC and JSC. 
\begin{table}[]
\footnotesize
\centering
\caption{Comparison with other methods on the left and right hippocampus segmentation.}
\label{tab:comparison}
\begin{tabular}{c|cccc}
\hline
     & DSC                                                                 & JSC                                                                 & PI                                                                  & RI                                                                  \\ \hline
NLP\cite{Coup2010Nonlocal}  & \begin{tabular}[c]{@{}l@{}}0.848/0.865\end{tabular} & \begin{tabular}[c]{@{}l@{}}0.752/0.764\end{tabular} & \begin{tabular}[c]{@{}l@{}}0.878/0.883\end{tabular} & \begin{tabular}[c]{@{}l@{}}0.857/0.864\end{tabular} \\
SPL\cite{Zhang2012Sparse}  & \begin{tabular}[c]{@{}l@{}}0.868/0.880\end{tabular} & \begin{tabular}[c]{@{}l@{}}0.763/0.778\end{tabular} & \begin{tabular}[c]{@{}l@{}}0.887/0.878\end{tabular} & \begin{tabular}[c]{@{}l@{}}0.866/0.875\end{tabular} \\
PBLF\cite{Wang2018Patch} & \begin{tabular}[c]{@{}l@{}}0.879/0.889\end{tabular} & \begin{tabular}[c]{@{}l@{}}0.773/0.789\end{tabular} & \begin{tabular}[c]{@{}l@{}}\textbf{0.903}/0.914\end{tabular} & \begin{tabular}[c]{@{}l@{}}0.879/\textbf{0.889}\end{tabular} \\
Ours & \begin{tabular}[c]{@{}l@{}}\textbf{0.897}/\textbf{0.900}\end{tabular} & \begin{tabular}[c]{@{}l@{}}\textbf{0.798}/\textbf{0.813}\end{tabular} & \begin{tabular}[c]{@{}l@{}}0.892/\textbf{0.919}\end{tabular} & \begin{tabular}[c]{@{}l@{}}\textbf{0.904}/0.882\end{tabular} \\ \hline
\end{tabular}
\end{table}
\section{Conclusion and Future Work}
In this work, we proposed a two-stage hippocampus
detection and segmentation framework based on 3D fully convolutional 
neural network, and we further improve the segmentation performance by introducing an enhancement mask. Our experiments demonstrate that
the enhancement mask generated from \textit{Proposal Network} speeds up the convergence of the training process and achieves significant improvements by concatenating with \textit{Segmentation Network}. 
Furthermore, our model can be easily extended to other biological tissue or organ detection and segmentation problems. 
As the future work, the two-stage neural network can be upgraded to an end-to-end trainable model, 
and the optimal value of parameters $\alpha$ and $\beta$ can also be learned from data.

% references section
% \begin{thebibliography}{1}

{\small
	\bibliographystyle{IEEEtran}
	\bibliography{crowd_analysis}
}
\end{document}